\documentclass[11pt]{article}

\usepackage[preprint]{acl}

\usepackage{times}
\usepackage{latexsym}
\usepackage{amsmath}
\usepackage{amsthm}
\usepackage{rotating}

\usepackage{booktabs}    
\usepackage{pifont}      
\usepackage{amssymb}     
\usepackage{xspace}
\usepackage{comment}
\usepackage{xcolor}
\usepackage{hyperref}

\usepackage{framed}
\usepackage{alltt}

\usepackage{listings}
\newcommand{\cmark}{\ding{51}}
\newcommand{\xmark}{\ding{55}}
\newcommand{\track}{\textsc{track}\xspace}

\usepackage[T1]{fontenc}

\usepackage[utf8]{inputenc}

\usepackage{microtype}

\usepackage{inconsolata}

\usepackage{graphicx}

\title{\track: Tool-Augmented Reasoning over Time Series\\ for Fault Diagnosis in Power Management Systems}

\author{Erkan Bayram, Debanjan Ghosh, Li Zhu, Sambarta Dasgupta \\
        Lorenz Labs,  Analog Devices Inc., USA \\
        \texttt{\{erkan.bayram,debanjan.ghosh,li.zhu,sambarta.dasgupta\}@analog.com}}

\begin{document}
\maketitle
\begin{abstract}
Fault diagnosis in power management systems running multiple loads relies on experts manually inspecting current and voltage waveforms, which fails to scale across deployed units. We present \track, which pairs a time series encoder with a Small Language Model to produce structured, physics-grounded diagnostic reports, calling tools for exact, verifiable calculations. It operates on-device, keeping proprietary sensor data local and avoiding expensive API calls. We introduce a rubric-based framework that evaluates these reports by scoring structure, signal fidelity, physics reasoning, diagnosis, and actionability through per-criterion verdicts. On a 15-class multi-load simulated benchmark, \track achieves over 95\% fault-category accuracy and generalizes to out-of-distribution faults, whereas the best encoder-probing baseline reaches only 58\% accuracy.
\end{abstract}

\section{Introduction} \label{section:introduction}

Modern power management systems supply and protect multiple independent \emph{loads} (components that consume electrical energy and convert it into work) through central or zonal power management units that continuously senses voltage and current across each channel. From electric vehicles to industrial robotics and grid infrastructure, nearly every electromechanical system depends on power management, and a single undiagnosed fault can halt the entire system. When faults occur, such as an LED failure, connector degradation, or fan stall, diagnostic signatures emerge in the sensed waveforms. Interpreting these signatures requires applying circuit-level physics (e.g., Ohm's law, power dissipation) and reasoning across multiple channels to isolate the root cause. In practice, this diagnosis is often performed manually by domain experts, a time-intensive and costly process that fails to scale across large deployments.

Language models can reason over signals and generalize  to unseen
faults, but large models require cloud-scale resources and incur costly latency.
Electromechanical systems generate proprietary data with limited bandwidth yet
carry onboard compute, motivating fault diagnosis \emph{at the edge} rather
in the cloud. Thus, we ask: can a multimodal Small Language Model (SLM) track diagnostic signatures in sensed load voltages and currents, producing reasoning traces, like chain-of-thought explanations~\cite{cot2022}, that are both physically grounded and actionable?

The challenge is fourfold. First, the model must combine multivariate time series data with varying baseline values (reference measurements under normal operation) and run on-device under memory and compute constraints, so a SLM is required. Second, the generated reasoning must contain a series of \emph{correct} numerical calculations (e.g., current and voltage analysis), a traditionally difficult task for SLMs because of their fragile numerical representations \cite{bhattacharya2025investigating}. Third, the model must reason over deviation patterns and explain them, especially for out-of-distribution (OOD) faults. Finally, evaluating the quality of physics reasoning requires domain-specific criteria that standard Natural Language Processing (NLP) metrics, which rely on text overlap, fail to capture. 

We address these challenges with \track, a tool-augmented architecture, and a rubric-based evaluation framework. Our contributions are:

\begin{itemize}
    \item A multimodal encoder-decoder architecture that generates structured reasoning across \textbf{five load channels} with 15 categories, conditioned on circuit simulation time series.
\item  Integration of verified tool 
functions guaranteeing exact arithmetic in reasoning traces similar to \emph{symbolic reasoning}~\cite{sprague2025cot}.
    \item A \textbf{rubric-based evaluation framework} featuring binary MET/UNMET criteria decomposed across five categories (e.g., structure, signal accuracy, physics reasoning quality, diagnosis, actionability), for in-distribution (ID), out-of-distribution, and multi-load settings~\cite{autorubric2026}.
    \item Deployment of \track on an edge device along with a real power
management system, executing the full pipeline from sampled signals to
reasoning traces, locally.
\end{itemize}

On a 15-class multi-load benchmark, our best configuration achieves over 95\% classification accuracy on a test set that combines ID and OOD faults, whereas the best encoder
probing reaches only about 58\% on the same test set. This gap demonstrates that language-grounded tool-call reasoning substantially outperforms encoder features alone. Our model produces  physically grounded diagnostic steps in natural language, reducing reliance on manual analysis and offering a scalable path for fault resolution across multi-load power management deployments.

\begin{figure*}[ht]
    \centering
    \includegraphics[width=0.95\linewidth]{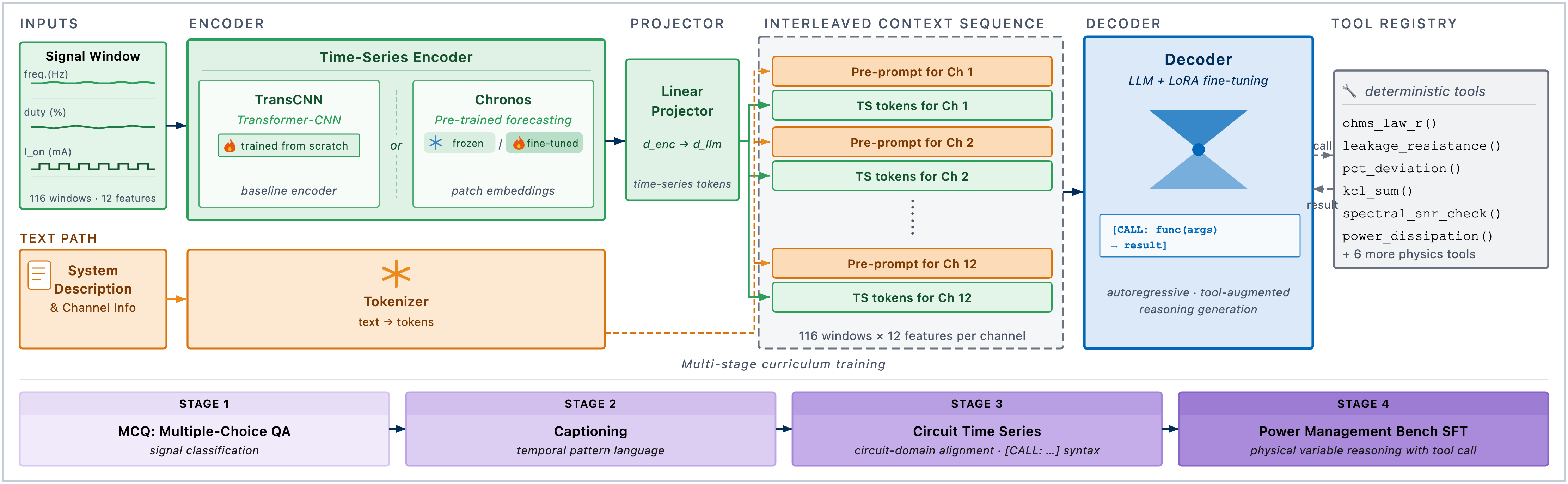}
    \caption{\track architecture. The Transformer-CNN encoder processes 12 sensor channels per load; projected embeddings are interleaved
  with text tokens (including per-load baseline references) and fed to the frozen Qwen~2.5-3B with LoRA. The tool executor intercepts
  \texttt{[CALL:]} patterns and injects exact results via named variables. Shades of purple indicate curriculum stages, with darker shades denoting later stages approaching final SFT.}
    \label{fig:arch_system}
\end{figure*}

\section{Datasets} \label{section:data}

\subsection{Time Series Data} \label{subsection:tsdata}

Existing time series reasoning datasets cover financial, weather, and physiological signals \cite{langer2025opentslm,yu2026tsrbench}, which differ from electrical circuit waveforms, and focus on classification, MCQ, or captioning. Fault diagnosis in
power management systems, however, requires data that tests generalization to
unseen faults and varying loads. Thus, we introduce a new time series dataset, called \textbf{Power Management Bench}, that is built around a simulated central power management unit driving five loads, where each load represents a different downstream component. They are (1) resistive, (2) thermal pad heating element, (3) Pulse-width modulated (PWM) resistive, (4) brushless Fan, and (5) LED strip. Using LTSpice, a free circuit simulation software, we simulate each configuration under varying noise levels~\cite{ltspice2026} to simulate faulty and healthy conditions across the loads, where a \textit{fault} is any anomalous condition that causes one or more signals to deviate from their nominal behavior. To ensure training distribution diversity, we vary sample counts across loads (e.g., 5,600 healthy samples for LED strip vs. 2,800 for Thermal pad). Faulty samples comprise approximately 25\% of the total dataset, reflecting the class-imbalance typical of real-world deployments. In total, 15 distinct categories including 5 healthy baselines and 10 faulty conditions are introduced (see Appendix, Table~\ref{tab:faults}). The simulation experiments produced 12 channels of 116 point time series, including on-state current $I_\text{ON}$, path resistance $R_\text{PATH}$, and other signals capturing voltage, noise, and spectral properties (see Appendix~\ref{sec:appendix_datasets} and Figure~\ref{fig:signals} for the simulation details).

\subsection{Reasoning Generation} \label{subsection:reason}
Next, we generate reasoning traces for each 12-channel time series for all the loads in a multi-stage pipeline: \textbf{(1)} An LLM agent receives fault signatures (e.g., insulation, LED short, connector
degradation), baseline values, hardware specifications via a custom MCP server, and human-authored reasoning 
examples for all the faulty and healthy categories. \textbf{(2)}  We apply best-of-3 rejection 
filtering: generate 3 reasoning candidates and then select the 
best (via a separate agent) based on completeness, clarity, and 
technical depth. \textbf{(3)} Reasoning traces are structured into
four sections as an expert conducts: \emph{signal profile} (raw measured value and time series motifs), \emph{system 
dynamics} (physics reasoning), \emph{source diagnosis} (i.e., fault severity), 
and \emph{summary} (classification and key finding) \textbf{(4)} We prompt the agent to employ Toulmin's argumentation scheme (i.e., claim, warrant, qualifier, rebuttal \newcite{toulmin1958uses}) so that each reasoning trace evaluates candidate fault hypotheses and provides rebuttals to rule out alternatives. \textbf{(5)} Training data embeds inline tool calls (i.e., ohms\_law\_r (i, v)->\$result)) with named variables e.g., \$result. Subsequent calls refer to {\$result} to prevent numerical re-typing. During training, tool results are pre-computed and the observation tokens are masked so the model trains on \emph{when to call} tools and \emph{which arguments to pass}, rather than memorizing \emph{results that the tool returns}. At inference, an executor intercepts each call, executes the function, and injects  the exact result(s), eliminating cascading arithmetic errors.

In addition, load specific tools are implemented, such as $\texttt{compute\_segments}$ to count faulty LED segments, and $\texttt{rise\_time\_deviation}$ to check resistive load. Finally, we diversify reasoning structure, tool call ordering, and vocabulary (via paraphrasing) and manually verify around 5\% of the data for correctness. Details on the structure and reasoning examples are in Appendix~\ref{sec:appendix_datasets} and Figure~\ref{fig:cot_comp}. 

\section{Experimental Details} \label{section:experiment}

\textbf{Model Architecture:} Our model follows the OpenTSLM soft-prompt architecture~\cite{langer2025opentslm}, coupling a time series encoder with a language model through a learned projector. The architecture as presented in Figure \ref{fig:arch_system} has three components: (a) a \textbf{Transformer-CNN encoder} (for brevity, Trans-CNN onwards; 6 layers, 8 heads, $d{=}128$, patch size 4), (b) a \textbf{linear projector} mapping from 128-dim to the LLM's 2048-dim, and (c) a \textbf{frozen Qwen~2.5-3B} \cite{qwen2025qwen25technicalreport} with LoRA~\citep{hu2022lora} ($r{=}16$, $\alpha{=}32$) that generates structured reasoning traces from projected sensor tokens. We experiment with two encoder initializations: (a) random (for Trans-CNN), and (b) warm-started from Chronos-T5-tiny~\cite{ansari2025chronos} (see Appendix~\ref{sec:appendix_model}). Each channel is preceded by a text pre-prompt of the form: \textit{``Load channel:{DC fan}, measured mean: 142.3\,mA, baseline: 140.0\,mA''}, providing the model with the channel identity and healthy-condition reference prior to processing the corresponding soft token embeddings. The language model receives the pre-prompts and soft prompt tokens of all channels interleaved in its context window.

\textbf{Model Training:} In both architectures, the projector is randomly initialized, so no shared grounding exists between the encoder output space and the language model's token space. Thus, directly applying supervised fine-tuning under these conditions risks incoherent reasoning. We therefore train through curriculum stages similar to OpenTSLM with some modifications. We inherit OpenTSLM's \emph{Stage 1} (MCQ: multiple choice QA) and \emph{Stage 2} (Captioning: description of M4 benchmark time series~\cite{m4}). However, neither dataset contains the multivariate electrical  signals (e.g., current waveforms, voltage droop, path resistance) that characterize our target domain. Both stages cover financial, weather, and business time series instead. Thus, we introduce a new time series dataset and corresponding simple reasoning structure after Stage 2 and before the final fine-tuning stage. This is our \emph{Stage 3} of curriculum training that we introduced, which is based on time series extracted from electrical circuits (the \emph{Circuit Time Series} (CTS) stage in Figure \ref{fig:arch_system}). CTS is a large-scale corpus of transient simulations from 1,000 diverse SPICE netlists (i.e., a plain-text description of an electronic circuit with simulations instructions). As mentioned, this stage exposes the model to the breadth of electrical signal behaviors before fine-tuning on the task-specific power management bench data (details in Appendix~\ref{appx:sec_cts} and Figure~\ref{fig:cot_st3}). Similar to the tool-calling functionality of the SFT stage, we introduced a few simple-tool calls in this stage 3. Finally, in Stage~4, the model is fine-tuned on the Power Management Bench dataset (Section~\ref{subsection:tsdata}) for fault diagnosis across five load channels and 15 categories including 10 faulty conditions, with class-weighted loss to address the natural imbalance between healthy and fault samples (see Section~\ref{subsec:training_setup} for training details). We use a total of 22K training examples, 2.75K validation and test examples, each. Additionally, to test performance beyond the training distribution, we also evaluate on OOD set of 1,050 samples with physically plausible failure modes across the loads (see Tables~\ref{tab:faults} and~\ref{tab:ood} in Appendix~\ref{sec:appendix_datasets})

\section{Rubric-Based Evaluation}
\label{sec:eval}

Standard NLP metrics such as ROUGE \cite{lin-2004-rouge} and BERTScore \cite{zhang2020bertscore} fail to discriminate between models in constrained technical domains. Since these metrics rely primarily on textual overlap, they suffer from numerical insensitivity and fail to penalize hallucination in numerical generation \cite{janiak-etal-2025-illusion,huang2025finnueexposingrisksusing}. Likewise, numeric LLM-as-judge scores (``rate from 0 to 1'') are poorly calibrated, brittle and irreproducible. Thus, we repurposed the binary rubric based evaluation as presented in AutoRubric~\cite{autorubric2026}. AutoRubric decomposes evaluation into multiple \textbf{binary MET/UNMET criteria} and demonstrated that binary criteria outperform numeric scores with better correlation with human judgments. We adapt their approach to a physics domain with three additions: (1)~code-verified criteria for numerical accuracy that do not require an LLM judge, (2)~domain-specific negative scoring for critical failures (wrong diagnosis: $-1$), and (3)~per-load criterion.

We organize evaluation into a total of 22 \textbf{binary MET/UNMET criteria} organized in five categories (see Table~\ref{tab:rubric} for rubric details). (a) \textbf{Structure}: (e.g., Valid JSON) (b) \textbf{Signal Profile}: (e.g., matches variables like IIN\_ON, etc.), (c) \textbf{Physics Reasoning}: (e.g., Ohm's law  check), (d) \textbf{Diagnosis}: (e.g., correct fault label), and (e) \textbf{Summary}: (e.g., contains diagnosis). Each criterion is either \textbf{code-verified} (JSON parsing, keyword matching, numerical tolerance) or \textbf{LLM-judged} (binary question to a Claude Haiku 5). The headline score is a weighted aggregate: $S\!=\!\sum_i w_i \cdot s_i / \sum_i w_i$, where $w_i$ reflects importance (e.g., correct fault label gets weight 2.0; physics reasoning gets 1.5) and $s_i \in \{-1, 0, 1\}$. The rubric is parameterized per load channel to test the tool usage and consistency.

\section{Results and Discussion} \label{section:result}

We evaluate generated diagnostics along two axes: fault classification and
reasoning quality.  

\begin{figure*}
    \centering
    \includegraphics[width=0.97\linewidth]{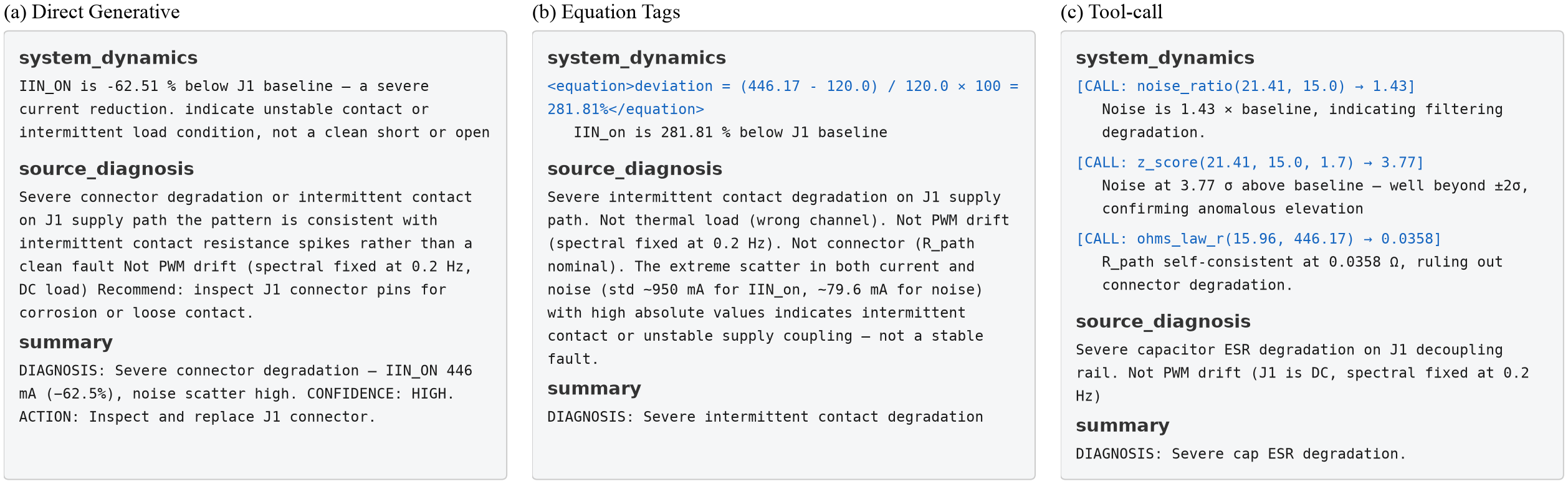}
    \vspace{-4mm}
  \caption{Reasoning trace comparison on a capacitive load. The direct generative (a) and equation tags (b)
  formats misdiagnose connector or contact degradation; only the tool-call format (c)
  computes the evidence with tools and correctly identifies capacitor ESR degradation.}
  \label{fig:cot_ood_example}    
\end{figure*}

\begin{table}[t]
\centering
\resizebox{\columnwidth}{!}{%
\begin{tabular}{lccc}
\toprule
\textbf{Model} & \textbf{ID} & \textbf{OOD} & \textbf{Combined} \\
\midrule
Trans-CNN + MLP                       & 45.53 & 26.10 & 40.16 \\
Chronos + MLP                         & 56.04 & 16.95 & 45.24 \\
\midrule
Trans-CNN + MLP with \textit{Baseline}          & 56.58 & {28.95} & 48.95 \\
Chronos + MLP with \textit{Baseline}            & {65.71} & 19.29 & 52.88 \\
\midrule
Trans-CNN + MLP with \textit{Baseline}  \&  \textit{Load} & 63.58 & {29.33} & {54.12} \\
Chronos + MLP with \textit{Baseline}  \&  \textit{Load}   & {72.22} & 20.29 & {57.87} \\
\bottomrule
\end{tabular}}
\caption{MLP classifier accuracy (\%) on encoder embeddings for
multi-load faults. \textbf{Combined} is weighted by ID and
OOD sample counts. \textit{Baseline} appends 11 scalar
references. \textit{Load} adds one-hot load encoding.}
\label{tab:baseline}
\end{table}

\paragraph{Non-reasoning Classifier}

To isolate the impact of the reasoning model, we train a lightweight Multi-Layer Perceptron (MLP) classifier on embeddings of the Stage~4-trained Trans-CNN or fine-tuned Chronos-T5-Tiny encoders. We progressively add the contextual inputs provided to the reasoning model in its pre-prompt: (i)~scalar reference measurements under healthy operating
conditions (\textit{Baseline}), and (ii)~a one-hot encoding of the load type (\textit{Load}). For OOD evaluation, we map each OOD fault in Table~\ref{tab:ood} to the nearest ID category based on severity
and load. Table~\ref{tab:baseline} shows that both inputs improve accuracy. While Chronos excels on ID samples, reaching 72.22\% against 63.58\%, Trans-CNN generalizes better to OOD samples (29.33\% vs. 20.29\%). We notice that domain-specific embeddings generalize better to unseen faults than general-purpose ones. Even so, the best probe reaches only 58\% on the combined test set, while the reasoning model achieves 95.42\% using the same inputs, confirming that the performance gap comes from reasoning.

\paragraph{Reasoning Model Evaluation}

We next discuss the proposed reasoning model in Table \ref{tab:training_configs}. For brevity, the table reports three metrics (see Table~\ref{tab:training_configs_full} for detailed version). \emph{Accuracy} is the average
fault-diagnosis accuracy across fault categories. \emph{Composite} is the
weighted rubric score $S$ of Section~\ref{sec:eval}, aggregated over binary
criteria covering signal-profile (criteria \#4-\#8), physics
reasoning (\#9-\#11), diagnosis (\#16-\#18), and recommended summary (\#19,\#21). \emph{Consistency} measures
if the conclusion follows logically from the reasoning body (\#22). Details on criteria (\#) in Table \ref{tab:rubric}.

\textbf{Effect of Reasoning Variants:} We compare three reasoning trace  variants: (1) \textbf{direct generative}: where the model estimates and calculates numbers in free text (like OpenTSLM~\cite{langer2025opentslm}), (2) \textbf{equation tags}: where the model writes formulas between equation tags but generates the results directly, thus the conclusion can disagree with the formula, and (3) \textbf{tool calls}: where calculations are offloaded to explicit functions and the returned value is chained into later steps, i.e., each number is \emph{computed} rather than \emph{generated} and can be \emph{verified} (see Figure~\ref{fig:cot_comp}). While all variants achieve similar 93--94\% accuracy, the composite scores reveal a clear hierarchy: 77.98\% (direct) vs. 85.85\% (equation tags) vs.  94.15\% (tool calls). We excluded tool-call related criteria and rescale the scores, so the first two variants are not penalized. However, even then, tool calls produce the most reliable reasoning, because only this variant guarantees correct numerical values and physics calculations.

On an OOD sample (Figure~\ref{fig:cot_ood_example}), the direct variant
hallucinates the deviation, while the equation-tag variant miscalculates it  (281.81\% instead of the correct value of 271.81\%). Both misclassify the fault. In contrast, 
 the tool-call variant computes the deviation via tool call and stores the result in a variable, later refers back to it and concludes with  the right diagnosis. This grounding in verifiable facts makes tool calls the most reliable for industrial deployment where unseen failures require traceable grounded reasoning.

\begin{table*}[h]
\centering
\small
\begin{tabular}{clcccccccc}
\toprule
\textbf{\#} & \textbf{Encoder} & \textbf{Calc. Style} & \textbf{St3\textsuperscript{a}} & \textbf{St4\textsuperscript{b}} & \textbf{Frozen\textsuperscript{c}} & \textbf{Accuracy} & \textbf{Consistency} & \textbf{Composite} \\
\midrule

    1 & Trans-CNN$_\text{BL}$     & Direct generative & \xmark & \cmark & \xmark & 92.70 & 95.52 & 77.98   \\
  2 & Trans-CNN$_\text{BL}$     & Equation tags     & \xmark & \cmark & \xmark & 93.88 & 96.06 & 85.85 \\
    3 & Trans-CNN$_\text{BL}$     & Tool call         & \xmark & \cmark & \xmark & \textbf{94.12} & \textbf{97.76} & \textbf{94.15} \\
\midrule
  4 & Chronos$_\text{F}$        & Tool call & \xmark & \cmark & \cmark & 92.70 & 96.82 & 94.09  \\
  5 & Chronos$_\text{Tr.}$      & Tool call & \xmark & \cmark & \xmark & 94.64 & 95.48 & 94.95 \\
6 & Chronos$_\text{F}$        & Tool call & \cmark & \cmark & \cmark & \textbf{95.42} & \textbf{97.97} & 95.01 \\
  7 & Chronos$_\text{Tr.}$      & Tool call & \cmark & \cmark & \xmark & {95.10} & {96.91} & \textbf{95.22} \\
  \midrule
    8 &Trans-CNN-CTS & Zero-shot (on St3)   & \cmark & \xmark & \xmark & 2.82 &   0.48 & 35.13 \\
  9 & Trans-CNN-CTS$_\text{F}$   & Tool call & \cmark & \cmark & \cmark & 94.21 & 96.58 & 93.89   \\
  10 & Trans-CNN-CTS$_\text{Tr.}$ & Tool call & \cmark & \cmark & \xmark & \textbf{94.61} & \textbf{97.45} & \textbf{95.12} \\
  
\bottomrule
\end{tabular}
\caption{Training configurations and calculation style ablation (scores in \%). Within each block,
best scores are \textbf{bold}.
\textsuperscript{a}St3: fine-tuning on the CTS dataset
with physics tool-call reasoning.
\textsuperscript{b}St4: fine-tuning for fault diagnosis on the Power Management Bench.
\textsuperscript{c}Frozen: encoder weights fixed during St4.
Composite is the rubric-based score of Section~\ref{sec:eval}.
Trans-CNN$_\text{BL}$ is randomly initialized, Trans-CNN-CTS is pretrained on
CTS. F\,=\,frozen, Tr.\,=\,trainable at St4.
\cmark denotes the presence of a training setup and \xmark ~denotes the absence. All configurations share
Stage~1 and~2.}\label{tab:training_configs}
\end{table*}

\textbf{Curriculum Learning and Encoder Effects} We first evaluate the zero-shot Stage~3 tool-call model, which measures how well the time series understanding learned on CTS transfers to fault diagnosis without Stage~4 supervision. Its composite score of 35.13\% shows that it already calls tools and produces structured reasoning while its low accuracy (2.82\%) highlights the need for domain alignment, which Stage~4 provides. However, Stage~3 familiarizes the model with electrical signals, and its tool-call reasoning further improves Stage~4 results, particularly benefiting frozen encoder configurations. For Chronos, Stage~3 improves accuracy from 92.70\% to 95.42\% with a frozen encoder, vs. only 94.64\% to 95.10\% with a trainable one, suggesting Stage~3 can substitute for encoder fine-tuning. With Stage~3, a frozen Chronos encoder outperforms all Trans-CNN-CTS variants in accuracy and consistency, reaching 95.42\% and 97.97\% against 94.61\% and 97.45\% for the better one. This is critical for industrial settings with limited labeled fault data. A frozen, off-the-shelf encoder avoids training from scratch and simplifies domain adaptation by fine-tuning only the projector and the language model. With proper tuning
of the surrounding stages, a frozen Chronos encoder is thus a
practical choice. While per-load performance is omitted for brevity, we observe that thermal pad, fan, and LED strip loads reach very high accuracy (in high 90\%) across all model variants, while PWM and resistive loads score between 63-78\%, due to the ambiguous boundaries between healthy baseline and early-stage faulty states. Calibrating such domain-specific fault understanding requires further collaboration with the power electronics community.

\textbf{Real-Time Deployment}\label{sec:deployment} We deploy \track on a Jetson Orin NX (16 GB) module on a live power management bench. We manually inject faults such as damaging insulation or stalling the fan;
since this provides too few samples for quantitative evaluation, we report
results on the simulated benchmark only (see Appendix~\ref{sec:appendix_demo}).
We also conduct a decoder scaling study across 0.5B, 1.5B, and 3B models
(Table~\ref{tab:model_size}). Accuracy stays at about 93\% for the 1.5B and
3B models and drops to 83\% for the 0.5B model, while consistency drops from 97.5\% to 59.3\% at 0.5B,
indicating that model size governs reasoning coherence more than label
prediction. Notably, the 0.5B tool-call model
outperforms the 3B direct-generative baseline (83.5\% vs 77.98\%),
confirming that structured reasoning ability contributes more to diagnostic
quality than decoder scale alone. Latency to generate scales from 42\,s to
65\,s across model sizes, offering a practical trade-off between reasoning
quality and response time.

\section{Related Work} \label{section:related}

\paragraph{Time-Series Reasoning.} Reasoning over time series has recently gained significant interest \cite{yang2026time,li-etal-2026-zara,messica2026adaptive}. SensorLM~\cite{sensorlm2025} aligns sensor and language at scale but produces only statistical captions without physics grounding. Other methods explore anomaly detection~\cite{yang2026time}, RL-based knowledge injection~\cite{sentsr2026}, human-in-the-loop feedback ~\cite{s2sfdd2026},  cognitive task taxonomy~\cite{tscognition2026}, and symbolic programs ~\cite{t2sp2026}. Closest to our work, OpenTSLM~\cite{langer2025opentslm} injects time series
patches into the embedding space of an LLM as soft prompts. Inspired by this design, \track differs by: (a) utilizing tool calls to ensure numerical stability,  physical correctness, and verifiability by construction and  
(b)~evaluating with a rubric that assesses reasoning
quality rather than accuracy alone. Time series foundation models such as Chronos~\cite{chronos2024} and
TimesFM~\cite{das2023decoder}, and LLM-based approaches such as
GPT4TS~\cite{zhou2023one} and TEMPO~\cite{cao2024tempo}, target forecasting,
whereas \track focuses on physics-aware reasoning.

\paragraph{Tool-Augmented Reasoning.} Following Toolformer's mid-generation token emission~\cite{toolformer2023} our architecture utilizes physics-specific tool calls. ToolRL~\cite{toolrl2025} showed RL with decomposed rewards outperforms SFT; we adopt SFT with observation masking and named-variable chaining as pilot RL experiments produced no additional gains. TimeClaw~\cite{timeclaw2026} employs an agentic runtime; our executor is simpler (10 stateless functions) but guarantees domain-specific correctness. KairosAgent~\cite{kairosagent2026} uses tool calls to form a textual morphology description in the prompt for a pretrained forecasting model whereas \track call physics-based tool during generation and applies them to conclude faults with a physics grounding.

\paragraph{LLM-as-Judge Evaluation.} Standard LLM judges with numeric scoring~\cite{zheng2023judging} are often poorly calibrated and irreproducible. AutoRubric~\cite{autorubric2026} achieves stable and interpretable assessments by decomposing evaluation into binary criteria. We adapt this to physics reasoning, checking for physical law application, self-consistency, and causal chain completeness. Appendix \ref{section:evalrubric} details our rubric.

\section{Conclusion and Future Work}

We present \track, a tool-augmented time series reasoning architecture for fault diagnosis in multi-load power management systems. By separating reasoning (employing physics law) from computation (executing exact arithmetic via tools), a 3B-parameter model achieves reliable diagnostics across 15 categories on five load channel. Our rubric-based evaluation provides actionable per-criterion feedback to guide targeted model improvement, a capability absent from standard NLP metrics. We compare different variants and observe off-the-shelf encoders (e.g., Chronos) seem a practical choice for real world deployments. Future work includes RL with tool-specific rewards~\cite{toolrl2025}, rubric-driven prompt optimization, and reducing latency in real-time implementation.

\section*{Limitations}

We have used a few generative tools for slight paraphrasing and grammar check in the text. SLM is used here to generate reasoning traces and certainly they have standard limitations such as non-deterministic nature of generation. We use large SOTA agents as LLM-as-a-judge, although, our binary rubric method is more effective, they can be brittle sometimes and hard to reproduce the exact accuracy values. Finally, although we have implemented our model running on a Jetson Orin module we plan to improve the compute efficiency and reduce the latency in the future. 

\section{Ethical Statement}

We do not foresee any potential negative societal impacts of our work. However, our reasoning model(s) are built upon LLMs, which are non-deterministic in nature. In reality, since our work is applied to Physical AI, extreme care must be taken before directly implementing LLM-generated outputs in real world environments.


\bibliography{custom}

\appendix

\section{Circuit Time Series (CTS)}\label{appx:sec_cts}

This dataset is built by running transient simulations on a collection of 1,000 open-source SPICE netlists~\cite{said2023circuit}\footnote{https://github.com/symbench/spice-datasets}. Each simulation produces a time series of node voltages (junctions in a circuit) and branch currents sampled at various sampling rate ranging between $10$ and $100$ kHz. We select these netlists such that they span a wide range of circuit types such as power regulators, oscillators, filters, and mixed-signal ICs. We only consider netlists with fewer than 15 nodes, which keeps the number of observed time series reasonable (i.e., fewer than 30). We also vary the settling time across circuits, so that the dataset covers the full range of transient behaviors.

\begin{figure*}[h]
    \centering
    \includegraphics[width=0.9\linewidth]{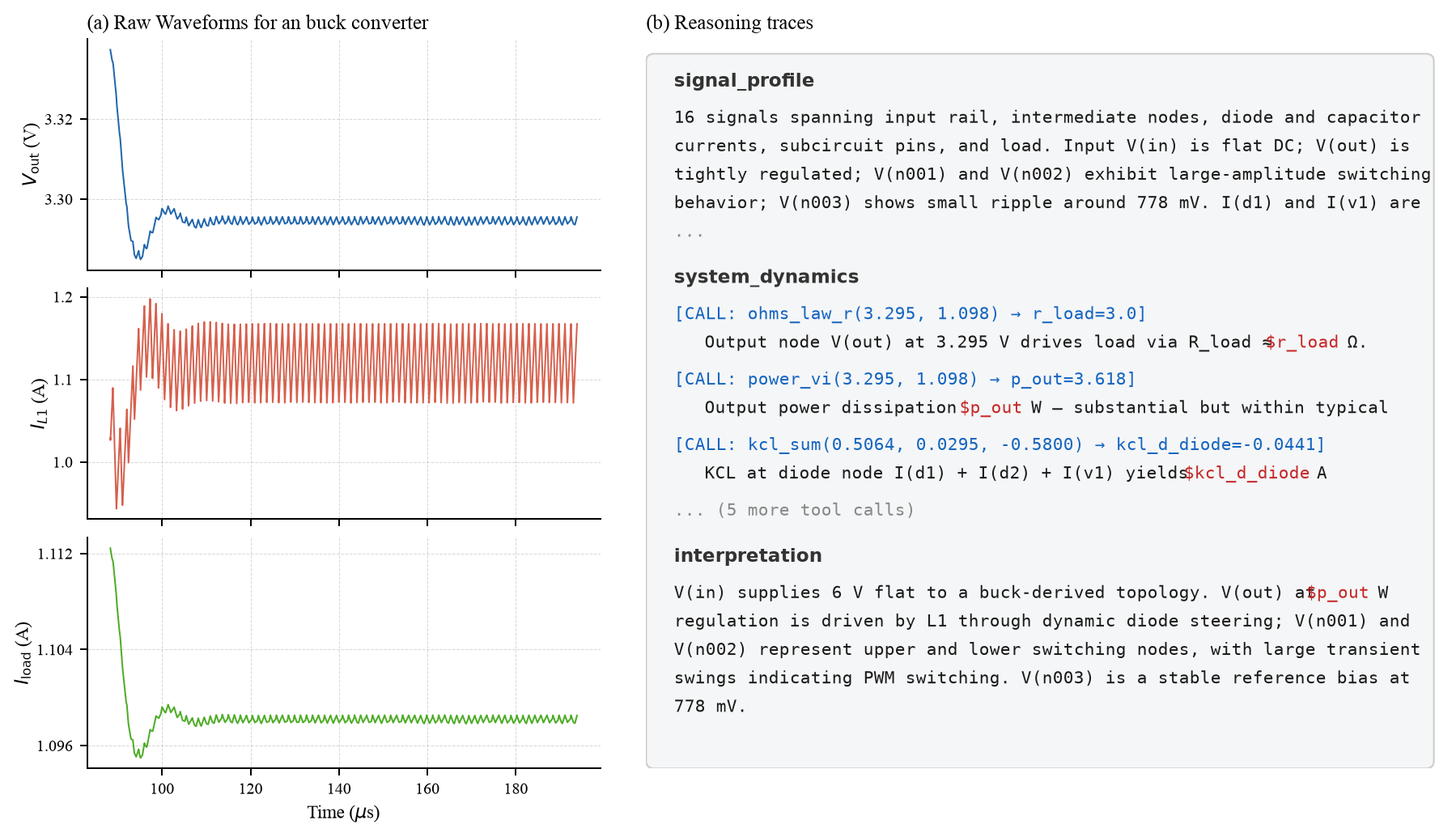}
    \caption{Example from the Circuit Time Series dataset. (a) Transient
waveforms from a buck converter circuit: output voltage $V_\mathrm{out}$,
inductor current $I_{L1}$, and load current $I_\mathrm{load}$.
(b) Structured reasoning traces showing physics tool calls
(blue) with chained variable references (red), followed by a natural
language interpretation. Tool calls invoke physics laws such as Ohm's
law, power calculation, and Kirchhoff's current law to ground the
reasoning in circuit physics.}
    \label{fig:cot_st3}
\end{figure*}

We then pair each time series with a structured annotation that contains explicit reasoning traces in the form of physics tool calls. For example, at a given node the annotation applies Kirchhoff's current law, which trains the model to verify the law rather than only describe the waveform. To the best of our knowledge, Circuit Time Series is the first large-scale time series corpus derived from circuit netlists and paired with a reasoning corpus (see Figure~\ref{fig:cot_st3} for an example).

\section{Power Management Bench Time Series}
\label{sec:appendix_datasets}

\begin{figure}
    \centering
    \includegraphics[width=0.95\linewidth]{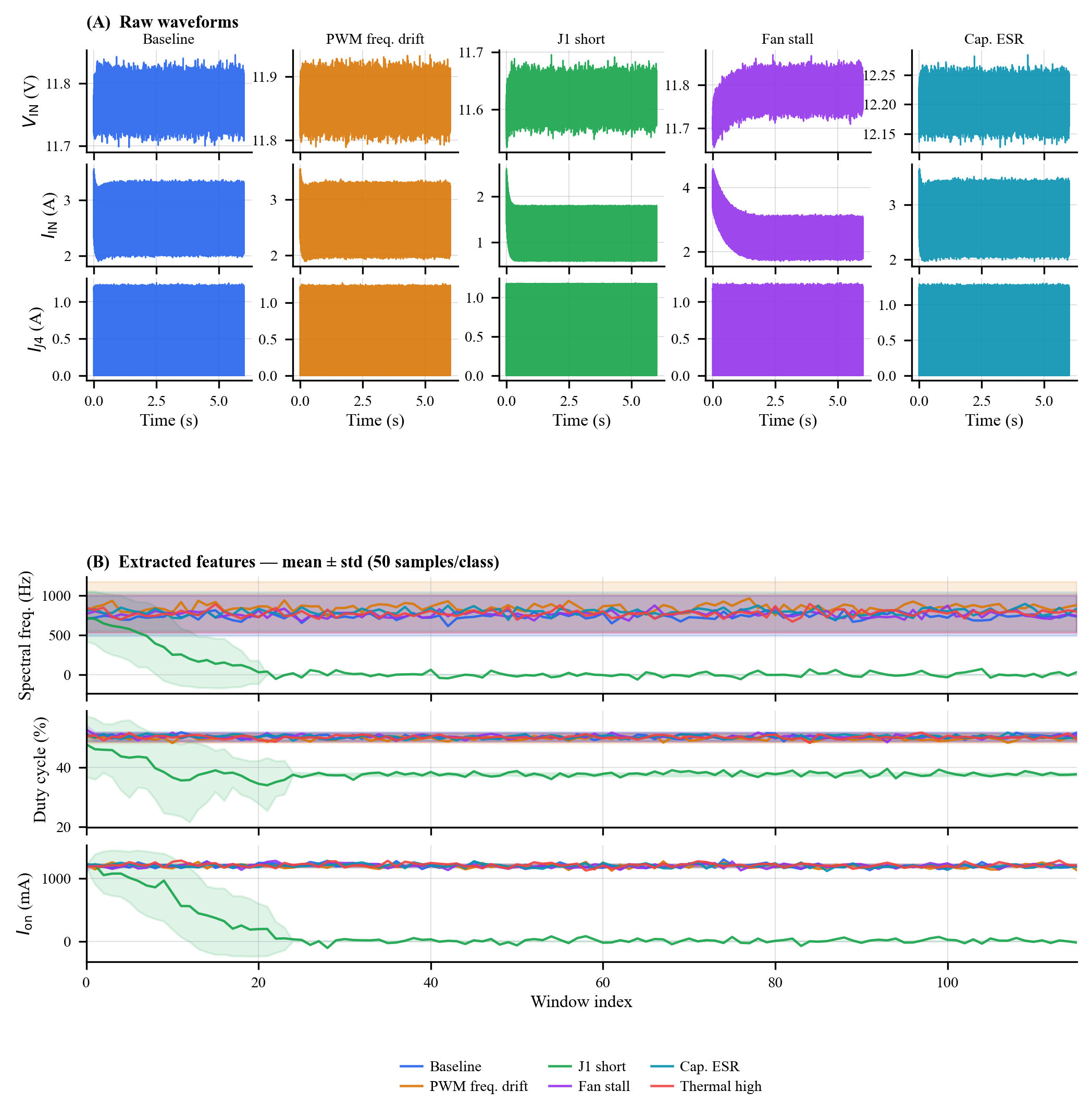}
    \caption{Power management bench signals and extracted per-window features across six different conditions.
(A) First $6$s of three synchronous channels ($V_\mathrm{IN}$, $I_\mathrm{IN}$, $I_{resistance}$) for five fault classes at $3200$ Hz.
(B) Mean $±$ 1 SD of three of extracted features over $116$ sliding windows ($50$ samples per class); shaded bands show within-class variability.}
    \label{fig:signals}

\end{figure}

We create a simulation of a power management system that drives five different loads. Similar to the Circuit Time Series dataset, we run the simulation in transient analysis and record raw voltage and current waveforms at 3.2kHz. We then process these raw waveforms with a sliding window (5\,s windows, 1\,s step, 120\,s recordings) to extract one scalar parameter per window per channel. This feature extraction yields a multivariate time series of length 116. Each window is characterized by 12 physical parameters, including on-state current $I_\text{ON}$, leakage current $I_\text{OFF}$, path resistance $R_\text{PATH}$, and nine others capturing voltage, noise, and spectral properties. Figure~\ref{fig:signals}(A) shows examples of the raw high-frequency waveforms. Figure~\ref{fig:signals}(B) shows how the extracted features separate the fault classes over the 116-window horizon.

For Power Management Bench time series, we expand the simulation on SPICE by  imitating the faults. Table \ref{tab:faults} presents the list of considered faults.  For example, we place a high-resistance leakage path from the branch conductor to ground to imitate insulation damage. Real insulation damage can be
intermittent and nonlinear, so we also add noise to the simulated signals to
approximate this variability. A detailed electrical model of such damage is
beyond the scope of this work.

\begin{table*}[h]
\centering
\caption{Dataset split statistics. OOD samples are mapped to the closest ID label for evaluation.}
\small
\begin{tabular}{llccc}
\toprule
Load Type & Condition & Train & Test & OOD \\
\midrule
LED strip & Healthy                 & 5{,}600 & 700 & 130 \\
          & Connector degradation   & 800     & 100 & 95  \\
          & Insulation damage       & 800     & 100 & 90  \\
          & LED short circuit       & 800     & 100 & 185 \\
\midrule
Resistive & Baseline (reference)    & 2{,}800 & 350 & --  \\
          & Cap.\ ESR (minor)       & 406     & 51  & 50  \\
          & Cap.\ ESR (severe)      & 394     & 49  & 50  \\
\midrule
Thermal   & Baseline (reference)    & 2{,}800 & 350 & 50  \\
          & Thermal overload (mod.) & 419     & 53  & 50  \\
          & Thermal overload (sev.) & 381     & 47  & 50  \\
\midrule
PWM       & Baseline (reference)    & 2{,}800 & 350 & 50  \\
          & Freq.\ drift (high)     & 400     & 50  & 50  \\
          & Freq.\ drift (low)      & 400     & 50  & 50  \\
\midrule
Fan       & Baseline (reference)    & 2{,}800 & 350 & --  \\
          & Fan stall               & 400     & 50  & 150 \\
\midrule
\multicolumn{2}{l}{Total} & 22{,}000 & 2{,}750 & 1{,}050 \\
\bottomrule
\end{tabular}
\label{tab:faults}
\end{table*}

\begin{figure*}
    \centering
    \includegraphics[width=\linewidth]{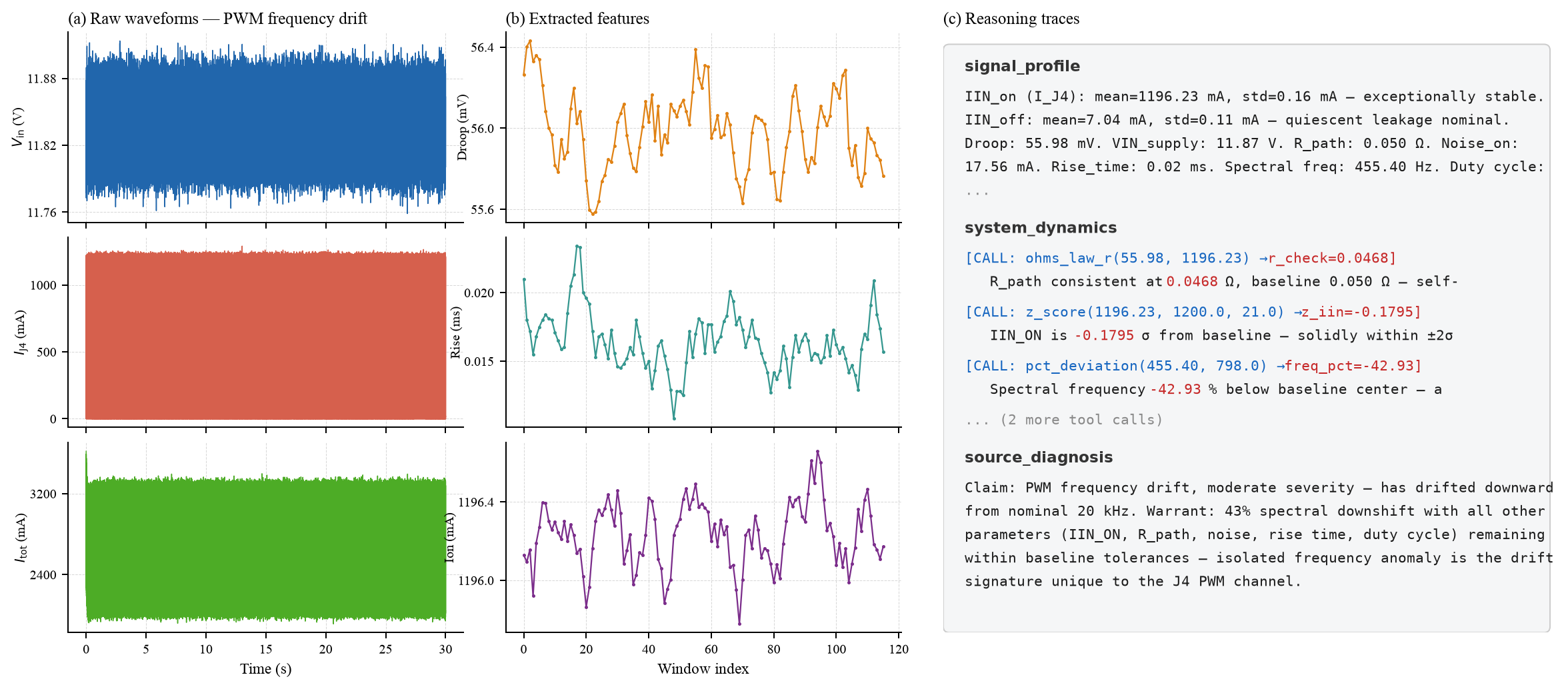}
\caption{Example from the Power Management Bench dataset illustrating
a PWM frequency drift fault on a load.
(a) Raw waveforms sampled at 3.2\,kHz: supply voltage $V_\mathrm{in}$,
PWM channel current $I_\mathrm{J4}$, and total input current
$I_\mathrm{total}$.
(b) Extracted per-window features: voltage droop, rise time, and on-state
current.
(c) Structured chain-of-thought annotation showing physics tool calls
(blue) with chained variable references, followed by a source diagnosis
grounding the fault in circuit physics.}
\label{fig:cot_st4}
\end{figure*}

Figure~\ref{fig:cot_st4} shows an example of the raw waveforms (a), the extracted per-window features (b), and the tool-augmented reasoning trace
(c) for a PWM frequency drift fault. The raw waveforms are noisy and are recorded at 3.2\,kHz. A six-second recording contains $1.92\times10^{4}$ samples per channel,
which yields $4.8\times10^{3}$ tokens per channel for the Trans-CNN encoder
(patch size four) and $1.2\times10^{3}$ tokens for the Chronos encoder after
patch and average pooling. Such a long input increases training and inference cost on an edge device and carries the measurement noise directly into the language model. We therefore select per-window features, such as voltage droop, rise time, and on-state current, that give the input a compact and interpretable
structure. These features suppress noise, expose the fault signatures directly, and let the model explain its diagnosis and apply physics tool calls to named quantities. Their extraction is lightweight enough to run on a small microcontroller, so it does not add to the computational burden of the language model. We use the features as model input. The raw waveforms are presented in Figure~\ref{fig:cot_st4} to help reader to cover the problem end to end. 


\begin{figure*}
    \centering
    \includegraphics[width=\linewidth]{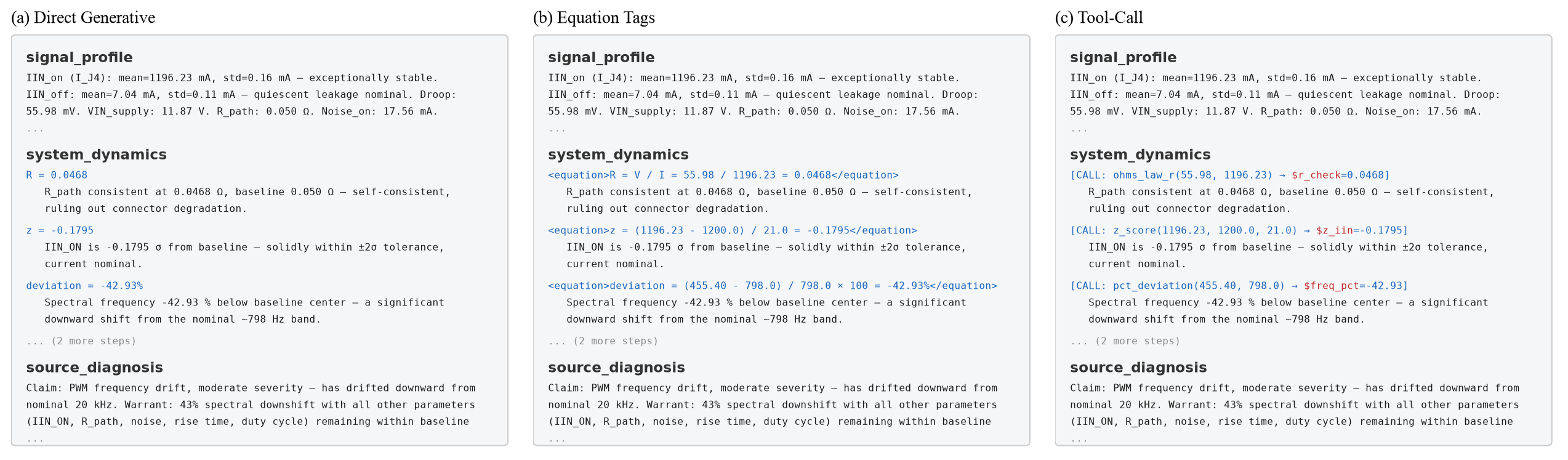}
     \caption{Comparison of reasoning traces formats for the same sample.}
    \label{fig:cot_comp}
\end{figure*}

Figure~\ref{fig:cot_comp} shows the three reasoning-trace variants in detail for the same
time-series sample, with each section annotated. In \texttt{signal\_profile}, the model
reports numerical statistics (e.g., mean and standard deviation) of the relevant features.
In \texttt{system\_dynamics}, it evaluates the individual system properties from the
provided load characteristics. For the PWM resistive load in this example, it computes the
path resistance and measures the deviation of the input current and spectral frequency
from their baselines (for a capacitive load, it would compute the impedance instead).
In \texttt{source\_diagnosis}, the model combines these two sections to reason about the
fault, stating a claim and linking it to the computed evidence that supports it.
This structure follows how an expert electrical engineer approaches fault detection and
diagnosis: characterize the signals, analyze the system with physics-based calculations,
and only then reason about the fault source.

\paragraph{OOD Faults:} To test model performance beyond the training distribution, we additionally construct an explicit out-of-distribution (OOD) set of 1050 samples with physically plausible failure modes intentionally selected to differ from the training fault conditions. Each OOD condition is mapped to its nearest ID label, the fault whose circuit behavior is most similar in terms of load characteristics and affected channel. The key challenge is that while the load and fault label are the same, the time series signatures differ: for example, a wrong-value resistor on the resistive load shares the same channel as a resistive short but produces a different current profile. The model is therefore expected to reason over the deviation pattern rather than recognize a memorized signature, correctly assigning the nearest label while identifying the true underlying anomaly.

\begin{table}
\centering
\small
\caption{OOD conditions and their affected loads.}
\label{tab:ood}
\begin{tabular}{lc}
\toprule
\textbf{OOD Condition} & \textbf{Load} \\
\midrule
Wrong resistor (5\,$\Omega$)      & Resistive load \\
Wrong resistor (20\,$\Omega$)        & Resistive load \\
Heatsink missing                    & Thermal pad    \\
Wrong pad type                    & Thermal pad    \\
Thermal near-short               & Thermal pad    \\
Duty cycle stuck high           & PWM load       \\
Duty cycle stuck low              & PWM load       \\
Wrong load (5\,$\Omega$)          & PWM load       \\
Fan high wattage                    & Brushless fan  \\
Fan low wattage                     & Brushless fan  \\
Fan bearing worn                    & Brushless fan  \\
More than $2$ LED is short in strip         & LED Driver \\
Higher current leakage ($20mA>$)    & LED Driver \\
Different number total LED in strip  & LED Driver \\
\bottomrule
\end{tabular}
\end{table}



\section{Model Architecture Details}
\label{sec:appendix_model}

\subsection{Transformer CNN:} We employ a encoder module (called Trans-CNN encoder). It splits the input time series into non-overlapping patches via a strided convolution and projects each patch into a fixed-dimensional embedding. We add learnable positional embeddings to encode patch position, apply layer normalisation and dropout, and pass the result to a stack of transformer encoder layers with multi-head self-attention and GELU feed-forward sublayers. Each channel is encoded independently, so the self-attention captures temporal structure within a single channel only. Cross-channel reasoning is delegated to the language model, which receives the soft prompt tokens of all channels interleaved with their text descriptions in its context window.

\subsection{Chronos Encoder:} We also employ Chronos-T5-Tiny~\cite{chronos2024} as the encoder backbone. In this case, each channel is encoded independently by the pretrained $T5$ encoder, which produces one hidden state $\mathbf{h}_t \in \mathbb{R}^{d_\text{T5}}$ per timestep. For channels exceeding Chronos's context window, the signal is split
into non-overlapping windows, each encoded independently, and the
resulting hidden states are concatenated along the temporal axis prior to
patch aggregation.
The hidden state sequence is then mapped to the soft prompt representation
through a two-stage aggregation procedure.

In the first stage, the $L$ per-timestep hidden states are partitioned into
$\lfloor L/p \rfloor$ non-overlapping patches of size $p{=}4$; within each
patch, the constituent vectors are concatenated along the feature axis to form
a $p \cdot d_\text{T5}$-dimensional representation, which is linearly
projected to the shared $128$-dimensional encoder output space and normalised. We consider concatenation to preserve the full within-patch temporal ordering, since short-duration transients such as switching events carry intra-patch diagnostic information that averaging suppresses.

In the second stage, the projected patch sequence of length $\lfloor L/p
\rfloor$ is further compressed by a factor of four via adaptive average
pooling to get a fixed-length soft prompt whose token count is independent
of the input series length. Overall, Trans-CNN encoder has $2.11$M parameters while Chronos-T5-Tiny has $4.33$M parameters, about twice as many.

\subsection{Training Setup.}\label{subsec:training_setup}
All models are trained on H100 GPUs using standard hyperparameters: AdamW
optimizer, bf16 precision, effective batch size of 16, and early stopping
with a patience of 5. Stages 1-3 are shared across configurations; St4
experiments branch from the same Stage~3 checkpoint, ensuring that
differences in final performance are attributable solely to St4 design
choices. For example, the frozen and trainable encoder variants of
Trans-CNN-CTS both initialize from the identical St3 checkpoint and
diverge only in whether the encoder weights receive gradients during St4
fine-tuning in the frozen variant only the projector and decoder are
updated, while in the trainable variant the full model is end-to-end
optimised.

\section{Real Time Deployment}\label{sec:appendix_demo}

We implement the system for real-time testing on a power management bench. Our
setup has three components: (i)~a power management bench that drives five loads
(resistive, thermal pad, PWM-controlled, fan, and an LED driver), (ii)~a
microcontroller board that samples the input voltage (VIN) and input current
(IIN) of these loads at 3.2\,kHz and extracts parameters from the samples, and
(iii)~a Jetson Orin module that runs the reasoning model for fault diagnosis.

We extract twelve parameters from a 5\,s sliding window that steps once per
second, such as on- and off-state current, voltage droop, and nine others for 2 minutes interval. The
model takes all twelve parameter series as input (total length of 116 for each). We deploy the Trans-CNN
encoder, linear projector, and SLM of Section~\ref{section:experiment} on the Jetson
module. The module runs capture, feature extraction, and inference in a single process and generates a reasoning trace with tool calls every 65\,s.

Table~\ref{tab:model_size} presents a decoder size study. We train the
Qwen~2.5~\cite{qwen2025qwen25technicalreport} decoder at three sizes, 0.5B, 1.5B, and 3B parameters, with the same
Trans-CNN-CTS encoder and the same four training stages. For each size, the
table reports accuracy, consistency, and composite score on the combined ID
and OOD test set, together with the inference latency per sample.

\begin{table}[h]
\centering
\small
\begin{tabular}{lcccc}
\toprule
\textbf{Size} & \textbf{Acc.} & \textbf{Consistency} & \textbf{Composite} & \textbf{Latency\,(s)} \\
\midrule
0.5B & 84.8 & 59.3 & 83.5 & $\sim$42$^*$ \\
1.5B & 93.3 & 75.8 & 88.8 & $\sim$53$^*$ \\
3B   & 94.6 & 97.5 & 95.1 & 65 \\
\bottomrule
\end{tabular}
\caption{Effect of decoder model size on rubric evaluation (scores in \%).
All configurations use the same Trans-CNN-CTS encoder architecture and they are trained through
St1-St4 with trainable encoder weights at St4. Latency is the inference time
per sample. The 3B model is measured directly on Jetson Orin NX (16GB),
and $^*$denotes expected run time for 0.5B and 1.5B scaled from their
H100 batch latency ratios relative to 3B.}
\label{tab:model_size}
\end{table}

\section{Individual Results}

We include the model performance on ID and OOD separately in the following Table~\ref{tab:training_configs_full}.

\begin{sidewaystable*}[t]
\centering
\small
\begin{tabular}{clccccccccccccc}
\toprule
 & & & & & & \multicolumn{3}{c}{\textbf{Accuracy}} & \multicolumn{3}{c}{\textbf{Consistency}} & \multicolumn{3}{c}{\textbf{Composite}} \\
\cmidrule(lr){7-9}\cmidrule(lr){10-12}\cmidrule(lr){13-15}
\textbf{\#} & \textbf{Encoder} & \textbf{Calc. Style} & \textbf{St3\textsuperscript{a}} & \textbf{St4\textsuperscript{b}} & \textbf{Frozen\textsuperscript{c}} & \textbf{ID} & \textbf{OOD} & \textbf{All} & \textbf{ID} & \textbf{OOD} & \textbf{All} & \textbf{ID} & \textbf{OOD} & \textbf{All} \\
\midrule
  1 & Trans-CNN$_\text{BL}$ & Direct generative & \xmark & \cmark & \xmark & 95.62 & 85.05 & 92.70 & \textbf{99.60} & 84.83 & 95.52 & 77.70 & 78.71 & 77.98 \\
  2 & Trans-CNN$_\text{BL}$ & Equation tags     & \xmark & \cmark & \xmark & \textbf{97.51} & 84.37 & 93.88 & \textbf{99.60} & 86.79 & 96.06 & 86.23 & 84.85 & 85.85 \\
  3 & Trans-CNN$_\text{BL}$ & Tool call         & \xmark & \cmark & \xmark & 97.05 & \textbf{86.45} & \textbf{94.12} & 99.49 & \textbf{93.23} & \textbf{97.76} & \textbf{93.93} & \textbf{94.73} & \textbf{94.15} \\
\midrule
  4 & Chronos$_\text{F}$   & Tool call & \xmark & \cmark & \cmark & 95.95 & 84.19 & 92.70 & \textbf{99.75} & 89.15 & 96.82 & 94.02 & 94.27 & 94.09 \\
  5 & Chronos$_\text{Tr.}$ & Tool call & \xmark & \cmark & \xmark & 96.85 & \textbf{88.85} & 94.64 & 99.35 & 85.34 & 95.48 & 95.11 & 94.53 & 94.95 \\
  6 & Chronos$_\text{F}$   & Tool call & \cmark & \cmark & \cmark & 98.07 & 88.48 & \textbf{95.42} & 99.27 & \textbf{94.57} & \textbf{97.97} & 95.01 & 95.01 & 95.01 \\
  7 & Chronos$_\text{Tr.}$ & Tool call & \cmark & \cmark & \xmark & \textbf{98.55} & 86.06 & 95.10 & 99.20 & 90.91 & 96.91 & \textbf{95.21} & \textbf{95.25} & \textbf{95.22} \\
\midrule
  8 & Trans-CNN-CTS & Zero-shot (St3) & \cmark & \xmark & \xmark & 3.20 & 1.82 & 2.82 & 0.58 & 0.22 & 0.48 & 35.71 & 33.61 & 35.13 \\
  9 & Trans-CNN-CTS$_\text{F}$   & Tool call & \cmark & \cmark & \cmark & 97.89 & 84.57 & 94.21 & \textbf{99.82} & 88.09 & 96.58 & 93.96 & 93.71 & 93.89 \\
 10 & Trans-CNN-CTS$_\text{Tr.}$ & Tool call & \cmark & \cmark & \xmark & \textbf{97.93} & \textbf{85.91} & \textbf{94.61} & 99.45 & \textbf{92.21} & \textbf{97.45} & \textbf{95.13} & \textbf{95.09} & \textbf{95.12} \\
\bottomrule
\end{tabular}
\caption{Training configurations and calculation style ablation, reported separately ID test set and OOD test set , and their combination (scores in \%).
All configurations are evaluated on the Power Management Bench test sets.
Best scores within each block are \textbf{bold}.
\textsuperscript{a}St3: fine-tuning on the CTS dataset with physics tool-call reasoning.
\textsuperscript{b}St4: fine-tuning on the Power Management Bench, the target task.
\textsuperscript{c}Frozen: encoder weights fixed during St4.
Composite is the rubric-based score of Section~\ref{sec:eval}.
Trans-CNN$_\text{BL}$ is randomly initialized, Trans-CNN-CTS is pretrained on CTS.
F\,=\,frozen, Tr.\,=\,trainable at St4.
\cmark\,=\,included, \xmark\,=\,not included.
All configurations share Stage~1 (MCQ) and Stage~2 (captioning) pre-training.}
\label{tab:training_configs_full}
\end{sidewaystable*}

\section{Evaluation Rubric} \label{section:evalrubric}

We present the rubric in Table~\ref{tab:rubric} that has been used to compute the composite score in Table~\ref{tab:training_configs} and Table~\ref{tab:training_configs_full}. On the one hand, most of the criteria are checked by code. For example, we verify that  the output have a valid JSON with all four sections, and it must stay within the word budget. The signal profile must report the measured values without interpretation and match the
ground-truth statistics. The physics reasoning must verify Ohm's law with approved tool calls
and ground its numbers in the prompt baseline with accurate equations. The diagnosis must match
the gold fault label, and the summary must contain a diagnosis, a confidence level, and an
action. On the other hand, an LLM judge covers the criteria that need semantic judgment (i.e. cross-parameter
reasoning, root-cause explanation, the causal chain from fault to signature, the rebuttal of
alternative faults). We also evalute whether  the conclusion follows the reasoning (reports as consistency metric). Some of the critical checks have also a penalty when they fail.

\begin{table*}
\small
\centering
\caption{Rubric: Evaluation Criteria}
\begin{tabular}{clp{8cm}lcc}
\toprule
\# & Category & Criterion & Method & Weight & Penalty \\
\midrule
1 & Structure       & All 4 sections present                          & code      & 1.0 & —    \\
2 & Structure       & Valid JSON                                       & code      & 1.0 & —    \\
3 & Structure       & Within word budget ($\leq$300; headlamp $\leq$450) & code    & 1.0 & —    \\
4 & Signal Profile  & No interpretive language in signal section       & code      & 1.0 & $-1$ \\
5 & Signal Profile  & $I_\text{IN,ON}$ value matches ground truth ($\pm$1\%)    & code      & —   & —    \\
6 & Signal Profile  & $I_\text{IN,OFF}$, Droop, $R_\text{path}$ match ($\pm$1\%, per load) & code & — & — \\
7 & Signal Profile  & Noise$_\text{ON}$ ($\pm$2\%; $I_\text{J1}$, $I_\text{FAN}$ only), rise time ($\pm$5\%; $I_\text{J1}$ only), spectral freq. ($\pm$5\%; $I_\text{J4}$ only) match & code & — & — \\
8 & Signal Profile  & Noise$_\text{OFF}$, duty, period, spectral SNR, $V_\text{IN}$ match & code & — & — \\
9 & Physics         & Ohm's law verified via tool call                 & code      & 1.5 & $-1$ \\
10 & Physics        & $\geq$2 tool calls present                       & code      & 1.0 & —    \\
11 & Physics        & All tool call names from approved registry       & code      & 1.0 & —    \\
12 & Physics        & Cross-parameter reasoning ($\geq$2 params linked) & LLM judge & 1.0 & —   \\
13 & Grounding      & Uses prompt baseline, not a memorized one        & code      & 0.5 & —    \\
14 & Grounding      & Equation accuracy (exact value match)            & code      & 1.5 & $-1$ \\
15 & Diagnosis      & Correct fault label (partial match OK)           & code      & 2.0 & $-1$ \\
16 & Diagnosis      & Root cause explained (physical mechanism)        & LLM judge & 1.0 & —    \\
17 & Diagnosis      & Fault $\rightarrow$ signature causal chain present & LLM judge & 1.0 & —  \\
18 & Diagnosis      & Rebuttal: rules out $\geq$1 alternative fault    & LLM judge & 1.0 & —    \\
19 & Summary        & Contains ``DIAGNOSIS:'' keyword                  & code      & 1.0 & —    \\
20 & Summary        & Contains confidence level (HIGH/MODERATE/LOW)    & code      & 1.0 & —    \\
21 & Summary        & Contains ``ACTION:'' keyword                     & code      & 1.0 & —    \\
22 & Consistency    & Conclusion follows reasoning body                & LLM judge & 1.5 & —    \\
\bottomrule
\end{tabular}
\label{tab:rubric}
\end{table*}

\end{document}